\documentclass[conference]{IEEEtran}
\IEEEoverridecommandlockouts
\usepackage{cite}
\usepackage{amsmath,amssymb,amsfonts}
\usepackage{algorithmic}
\usepackage{graphicx}
\usepackage{textcomp}
\usepackage{xcolor}
\usepackage{hyperref}
\usepackage{caption}
\usepackage{subcaption}
\usepackage{multirow}
\usepackage{xurl}
\usepackage{comment}

\def\BibTeX{{\rm B\kern-.05em{\sc i\kern-.025em b}\kern-.08em
    T\kern-.1667em\lower.7ex\hbox{E}\kern-.125emX}}
\begin{document}

\title{Smart Energy Guardian: A Hybrid Deep Learning Model for Detecting Fraudulent PV Generation
\thanks{*Corresponding author: Hao Wang (hao.wang2@monash.edu).}
\thanks{This work was supported in part by the Australian Research Council (ARC) Discovery Early Career Researcher Award (DECRA) under Grant DE230100046.}
}

\author{\IEEEauthorblockN{Xiaolu Chen\textsuperscript{1}, Chenghao Huang\textsuperscript{2}, Yanru Zhang\textsuperscript{1,3}, Hao Wang\textsuperscript{2,4,*}}
\IEEEauthorblockA{
\textsuperscript{1}School of Computer Science and Technology, University of Electronic Science and Technology of China, Chengdu, China \\
\textsuperscript{2}Department of Data Science and AI, Faculty of Information Technology, Monash University, Melbourne, Australia \\
\textsuperscript{3}Shenzhen Institute for Advanced Study, Shenzhen, China\\
\textsuperscript{4}Monash Energy Institute, Monash University, Melbourne, Australia
}
}

\IEEEoverridecommandlockouts
\IEEEpubid{\makebox[\columnwidth]{979-8-3503-6431-6/24/\$31.00 \copyright2024 IEEE \hfill} \hspace{\columnsep}\makebox[\columnwidth]{ }}

\maketitle

\IEEEpubidadjcol

\begin{abstract}
With the proliferation of smart grids, smart cities face growing challenges due to cyber-attacks and sophisticated electricity theft behaviors, particularly in residential photovoltaic (PV) generation systems. Traditional Electricity Theft Detection (ETD) methods often struggle to capture complex temporal dependencies and integrating multi-source data, limiting their effectiveness. In this work, we propose an efficient ETD method that accurately identifies fraudulent behaviors in residential PV generation, thus ensuring the supply-demand balance in smart cities. Our hybrid deep learning model, combining multi-scale Convolutional Neural Network (CNN), Long Short-Term Memory (LSTM), and Transformer, excels in capturing both short-term and long-term temporal dependencies. Additionally, we introduce a data embedding technique that seamlessly integrates time-series data with discrete temperature variables, enhancing detection robustness. Extensive simulation experiments using real-world data validate the effectiveness of our approach, demonstrating significant improvements in the accuracy of detecting sophisticated energy theft activities, thereby contributing to the stability and fairness of energy systems in smart cities.
\end{abstract}


\section{Introduction}

With the widespread deployment of smart grids, modern power systems are increasingly vulnerable to cyber-attacks and evolving electricity theft behaviors~\cite{9686052}. Traditionally, electricity theft involved physical methods, such as illegal access to the grid. However, in modern power systems, cyber-attacks have introduced new avenues for theft, enabling attackers to steal electricity by hacking smart meters, tampering with billing data, or compromising the IT infrastructure of power companies.

As the penetration of renewable energy sources continues to grow, a new form of electricity theft has emerged, particularly among users with distributed generation installations based on these sources. These prosumers may manipulate generation data to falsely report higher electricity production to power companies, thereby illicitly increasing their financial gains~\cite{8355252}. This fraudulent behavior not only results in direct financial losses for power companies but also threatens the stability and security of the power system. Such theft can lead to imbalances in power supply, potentially causing localized outages or even widespread grid failures~\cite{7438916}. Furthermore, electricity theft driven by cyber-attacks undermines the fairness of the electricity market, leading to increased prices for legitimate consumers and impacting social stability~\cite{en13174331}. Consequently, preventing electricity theft, especially in the context of cyber-attacks, has become an urgent priority to ensure the security of the power system and maintain social stability.

Significant progress has been made in electricity theft detection (ETD) in the consumption domain, where various methods have been successfully employed to identify theft activities~\cite{7108042, 9205903, 9909779, WANG2024123228}. However, as electricity theft increasingly shifts from the consumption domain to the generation domain, particularly among distributed photovoltaic (PV) customers, new challenges arise that necessitate more advanced detection approaches. In response to this shift, researchers have begun to focus on ETD in the generation domain, recognizing the growing importance of addressing theft in this area.
Typically, the data associated with normal electricity consumers dominate, while data related to malicious consumers are scarce. This imbalance often causes models to be biased toward the majority class during training, leading to a diminished ability to accurately identify theft behaviors. To address this challenge, researchers propose various technical approaches.
Li et al.~\cite{9712854} proposed a method for enhancing distributed photovoltaic electricity theft sample data using a Wasserstein generative adversarial network. This approach addresses the issue of insufficient theft samples, thereby improving the accuracy of electricity theft detection.
Shaaban et al.~\cite{9528289} developed an unsupervised anomaly detection system based solely on benign data, incorporating a variety of data sources. This data-driven approach, grounded in machine learning, is designed to capture the intrinsic characteristics of benign data and subsequently detect any deviations from this baseline.
In addition, to enhance the performance of detection systems, researchers not only focus on utilizing single data sources but also explore the integration of multiple data sources.
Maala et al.~\cite{10322383} investigated the performance of five features and four algorithms employed in theft detection models for distribution systems integrating rooftop solar panels and net metering.
Eddin et al.~\cite{9964082} proposed a deep learning-based cyber-attack detection model that utilizes a single data source. They further examine the robustness of the model against small perturbation attacks.
However, information from a single data source may be insufficient to fully capture the complexity and diversity of electricity theft. Therefore, some studies explored the utilization of multiple data sources to develop more comprehensive and robust detection models~\cite{9528289}.
Ismail et al.~\cite{8998142} proposed a set of cyber-attack functions to simulate the manner malicious consumers steal electricity by manipulating a renewable-generation smart meter, and developed a deep learning-based electricity theft detection system using various data sources to detect electricity theft cyber-attacks in solar panels.

Although existing research has made progress in addressing the challenge of improving detection performance, the limitations of traditional methods have become increasingly apparent with the growing volume of data and the rising complexity of theft behaviors. The two main challenges posed by previous works on PV generation theft detection are summarized as follows.
\begin{itemize}
    \item \textbf{Limited temporal dependency capture:} Traditional methods, including classical ML and basic DL models, struggle to capture complex temporal dependencies, leading to inaccurate detection of sophisticated, time-evolving fraudulent activities. These methods often miss subtle temporal patterns essential for identifying energy theft.
    \item \textbf{Inadequate multi-source data integration:} Traditional approaches often inadequately integrate multi-source data and external factors. This limits their ability to fully exploit contextual influences, like temperature variations, resulting in less robust detection.
\end{itemize}

The goal of this work is to develop an efficient ETD method that accurately identifies theft behaviors in residential PV generation, ensuring the supply-demand balance in smart cities, and thus enhancing the stability and fairness of smart city energy systems. To address the challenges of capturing complex temporal dependencies and integrating multi-source data, our method combines multiple types of neural networks into a hybrid model, exceling at identifying sophisticated, time-evolving fraudulent activities while effectively incorporating contextual factors like temperature variations.
The main contributions of our work can be summarized as follows.
\begin{itemize}
    \item \textbf{Hybrid DL model for dependency capture:} We develop a hybrid model combining multi-scale Convolutional Neural Network (CNN), Long Short-Term Memory (LSTM), and Transformer, that effectively captures both short-term and long-term temporal dependencies among multi-source time series, including residential load, PV generation, and daily patterns among all prosumers, significantly enhancing the accuracy of energy theft detection.
    \item \textbf{Modal fusion using data embedding}: Considering poor accessibility of detailed temperature, our approach achieves seamless integration of time-series data and discrete variables through data embedding, allowing for more accurate integration of external contextual factors, such as statistical temperature data, into the detection process.
    \item \textbf{Validation through real-world data simulation:} We conduct extensive simulation experiments using real-world data, demonstrating the effectiveness and robustness of our proposed method in accurately detecting energy theft.
\end{itemize}

\section{Problem Statement}
\subsection{Generation Theft}
\begin{figure}[t]
    \centering
    \includegraphics[width=0.95\linewidth]{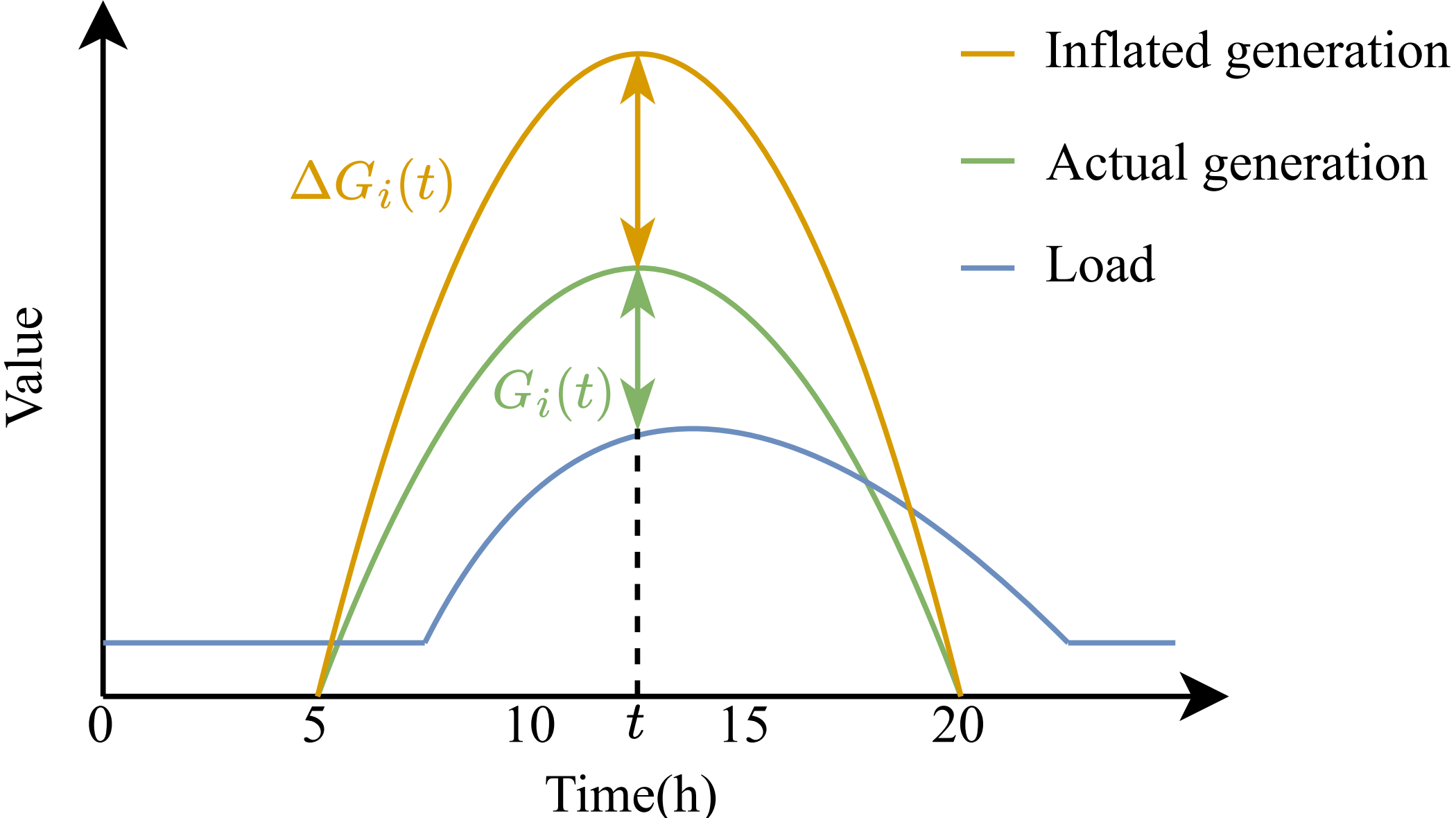}
    \caption{The overview of PV generation theft.}
    \label{fig:overview}
\end{figure}

Suppose a prosumer without battery will only probably carry out theft behaviors on generation, while the load recorded by smart meters is honest. We denote the actual load and generation at time $t$ on the $i$th day as $L_i(t)$ and $G_i(t)$, respectively. The prosumer reports their electricity generation to the power company as $\hat{G_i}(t)$. Under normal circumstances, the reported electricity generation should match the actual electricity generation, i.e. $\hat{G_i}(t)=G_i(t)$. However, in cases of electricity theft, the prosumer may report an inflated generation value to illegally gain extra profits. This behavior can be represented as
\begin{align}
    \hat{G_i}(t)=G_i(t)+\Delta G_i(t),
\end{align}
where $\Delta G_i(t)$ represents the inflated electricity generation reported by the prosumer, and $\Delta G_i(t)>0$ indicates that the prosumer is reporting a higher generation than the actual value to obtain excessive earnings as shown in Figure~\ref{fig:overview}. 
This fraudulent reporting causes the total electricity generation received by the power company to be overestimated.
Therefore, the objective is to design a detection model $D(\cdot)$ that can accurately identify instances of fraudulent generation reports by prosumer $i$, specifically detecting cases where $\Delta G_i(t)>0$, thereby maintaining the fairness of the electricity market and the stability of the power system. Note that, to prevent misjudgments caused by prediction errors of the original $G_i(t)$ or other physical factors leading to inaccurate readings, we set a threshold $\epsilon$ to tolerate minor discrepancies. For a given reported generation value $\hat{G_i}(t)$, the model output is defined as
\begin{align}
    D(\hat{G_i}(t))=\begin{cases} 
1, & \text{if }  \frac{\hat{G}_i(t)}{G_i(t)} > \epsilon, \\
0 ,& \text{if } \frac{\hat{G}_i(t)}{G_i(t)} \leq \epsilon,
\end{cases}
\end{align}
where $D(\hat{G_i}(t))=1$ indicates the detection of electricity theft, and $D(\hat{G_i}(t))=0$ indicates no electricity theft detected.

\subsection{Patterns of Load and Generation}
Furthermore, since the overall pattern of the reported electricity generation aggregated from a bundle of prosumers may provide additional insights to the power company, it is reasonable to set the patterns of all prosumers as a part of the input variables.
Assume the power company has $N$ prosumers, then the pattern of aggregated reported electricity generation at time $t$, recorded by the smart meters, can be obtained as follows
\begin{align}
    P^{\hat{G}}(t)=\frac{1}{N}\sum_{n=1}^{N}\hat{G_i^n}(t)=\frac{1}{N}\sum_{n=1}^{N} \big[G_i^n(t)+\Delta G_i^n(t)\big],
\end{align}
while the actual electricity load and generation patterns are
\begin{align}
    P_i^L(t)=\frac{1}{N}\sum_{n=1}^{N}L^n_i(t),\\
    P_i^G(t)=\frac{1}{N}\sum_{n=1}^{N}G^n_i(t).
\end{align}

Finally, since the goal is to detect whether any theft behavior exists in daily PV generation, we extend each of the four variables into a series over a horizon, denoted as $T$: $\{L_i(t)\}_{t=1}^T$, $\{\hat{G}_i(t)\}_{t=1}^T$, $\{P_i^{\hat{G}}(t)\}_{t=1}^T$, and $\{P_i^L(t)\}_{t=1}^T$.

\section{Methodology}

\begin{figure}[t]
    \centering
    \includegraphics[width=0.99\linewidth]{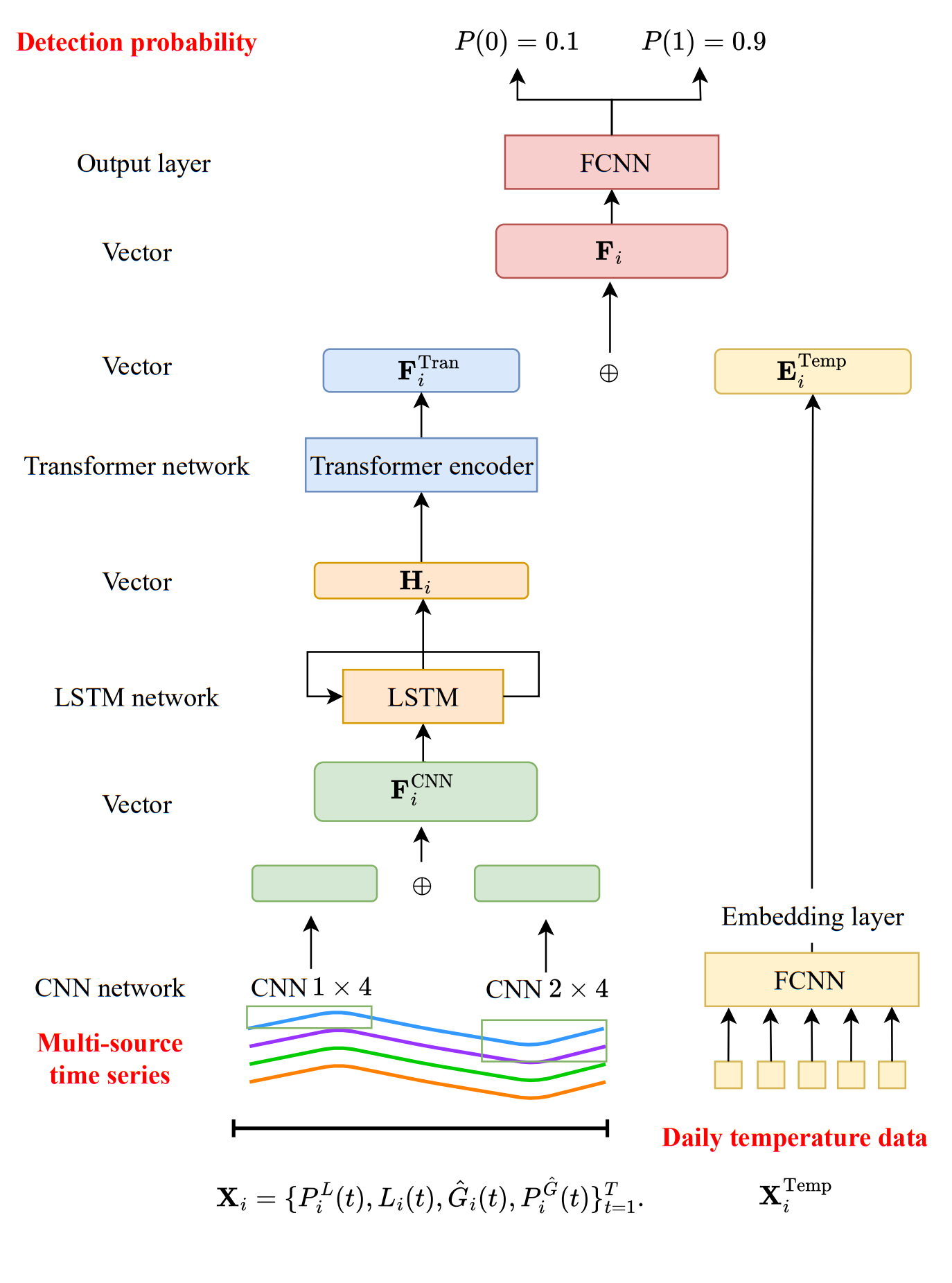}
    \caption{The overview architecture of the proposed framework.}
    \label{fig:method}
\end{figure}

In this work, we propose a novel approach for detecting energy theft in residential solar generation by leveraging multi-scale Convolutional Neural Network (CNN)~\cite{cui2016multi}, Long Short-Term Memory (LSTM)~\cite{graves2012long}, and Transformer~\cite{vaswani2017attention}. The goal is to accurately identify fraudulent behaviors where prosumers report inflated energy generation data to illegally boost their revenue. The proposed methodology integrates the strengths of CNN in extracting short-term features among variables, LSTM in capturing temporal auto-correlations, and Transformer in modeling complex relationships across different data sources. Additionally, an embedding mechanism is introduced to incorporate the prosumer's temperature data, including daily high, low, median, standard deviation values, and the corresponding season. The overview of the proposed method is illustrated in Figure~\ref{fig:method}.

\subsection{Architecture of Proposed DL Model}
In this section, we introduce the architecture of the proposed NN following the order of the data flow.

\subsubsection{Input Variables}

The detection model processes multi-source time series data, including the reported PV generation, actual load of one prosumer, and the load pattern and the generation pattern of all prosumers over a period on the $i$th day. Each data source contributes to the detection of potential anomalies indicative of energy theft.

Let $\mathbf{X}_i$ represent the multi-source input data for prosumer $i$ over the time window $T=24$, such that
\begin{align}
    \mathbf{X}_i = \{P_i^L(t), L_i(t), \hat{G}_i(t), P_i^{\hat{G}}(t)\}_{t=1}^T.
\end{align}
Thus, a total of four series are included in $\mathbf{X}_i$. Note that, for multi-scale CNN processing, the four series are arranged in a suitable order.

\subsubsection{Multi-Scale CNN for Short-Term Dependencies}
First, the input variables are processed through a multi-scale CNN to capture short-term temporal dependencies of both univariate series and mutli-variate series. Specifically, convolutional filters with various kernel sizes are applied.

The kernel with $1\times4$ size is applied to all four series to capture local short-term dependencies and patterns, such as abrupt changes in PV generation, which may indicate anomalous behaviors. This kernel effectively looks at a four-hour segment at a time, identifying fine-grained temporal patterns.

On the other hand, the kernel with $2\times4$ size is applied to detect slightly broader trends and correlations between consecutive data points, which are crucial for identifying gradual yet systematic anomalies in the data.

The output of these convolutions produces a set of feature maps $\mathbf{F}^{\text{CNN}}_i$, where each map corresponds to the different temporal patterns captured by the various kernel sizes
\begin{align}
    \mathbf{F}^{\text{CNN}}_i = \big[ \text{CNN}_{1\times4}(\mathbf{X}_i), \text{CNN}_{2\times4}(\mathbf{X}_i)\big],
\end{align}
where $\text{CNN}_{k\times4}$ represents the convolution operation with a kernel size of $k\times4$.

After convolution, the feature maps are passed through a max-pooling layer to reduce their dimensionality and enhance the most salient features, preparing them for subsequent processing by the LSTM network.

\subsubsection{LSTM for Long-Term Dependencies}

The CNN-extracted features $\mathbf{F}^{\text{CNN}}_i$ are passed through an LSTM network, which models the temporal dependencies in the time series data. The LSTM outputs the hidden state $\mathbf{H}_i$ that captures sequential patterns across the 24-hour window as
\begin{align}
    \mathbf{H}_i = \text{LSTM}(\mathbf{F}^{\text{CNN}}_i; \theta^{\text{LSTM}}),
\end{align}
where $\theta^{\text{LSTM}}$ represents the LSTM network's parameters.

\subsubsection{Transformer for Correlations Between Series}
To enhance the model's ability to model relationships among multiple data sources across the 24-hour period, the LSTM hidden states $\mathbf{H}_i$ are further processed by a Transformer encoder. The Transformer provides a global context, enabling the model to detect subtle and complex patterns indicative of energy theft as
\begin{align}
    \mathbf{F}^{\text{Tran}}_i = \text{Transformer}(\mathbf{H}_i; \theta^{\text{Tran}}),
\end{align} 
where $\theta^{\text{Tran}}$ are the parameters of the Transformer.

\subsubsection{Embedding for Temperature Data}
Since the temperature data consists of only five statistical values in one day, a fully connected neural network (FCNN) is employed to embed this information into a vector of fixed size as
\begin{align}
    \mathbf{E}_i^{\text{Temp}} = \text{FCNN}(\mathbf{X}_i^{\text{Temp}}; \theta^{\text{Temp}}),
\end{align}
where $\mathbf{X}_i^{\text{Temp}}$ are the statistical temperature values on the $i$th day, and $\theta^{\text{Temp}}$ is the parameters of the FCNN for temperature embedding.

\subsubsection{Detection Probability Output}
The temperature embedding $\mathbf{E}_i^{\text{Temp}}$ is concatenated with the output from the Transformer $\mathbf{F}^{\text{Tran}}_i$, forming the combined feature vector as
\begin{align}
    \mathbf{F}_i = [\mathbf{F}^{\text{Tran}}_i, \mathbf{E}_i^{\text{Temp}}].
\end{align}
Finally, this vector is passed through a FCNN to output the probability of detecting energy theft, which is presented as
\begin{align}
    P\big[D\big(\hat{G}_i(t)\big)=1\big] = \text{Softmax}(\mathbf{F}_i; \theta^{\text{FCNN}}),
\end{align}
where $\theta^{\text{FCNN}}$ is the parameters of the FCNN for final output.

\subsection{Model Training}
Denote $P\big[D\big(\hat{G}_i(t)\big)=1\big]$ as $\hat{P}_i$. The model is trained using a cross-entropy loss function, defined as
\begin{align}
    \mathcal{L} = -\mathbb{E}\big[y_i\log \hat{P}_i + (1-y_i)\log (1-\hat{P}_i) \big], i \in \{1,...,N\},
\end{align}
where $y_i$ is the ground truth label indicating the presence (1) or absence (0) of energy theft.

During training, the model learns to minimize this loss, optimizing the parameters of the CNN, LSTM, Transformer, and the temperature embedding network. This comprehensive training process enables the model to effectively detect energy theft by capturing both temporal patterns among multi-source time series data and the impact of daily temperature variations.

\section{Evaluations}
In this section, we provide a detailed description of the datasets and evaluation metrics employed in our study, along with an introduction of the benchmark methods selected for our experiments. 
Subsequently, a comprehensive analysis of the performance of the benchmark methods and the proposed method is performed, including a comparison of the effectiveness of each method on various evaluation metrics.

\subsection{Data setup}
The dataset employed in this study is a synthetic dataset of Danish residential electricity prosumers~\cite{yuan2023synthetic}. The dataset is a comprehensive dataset containing import and export electricity data from five categories of residential prosumers. Specifically, the dataset includes solar export data, detailing the hourly electricity output from PV systems installed by individual residential prosumers, which is subsequently fed into the grid. The data are annotated with date type, season, and daily temperature, comprising $50,000$ days of solar generation records. Therefore, it is reasonable to assume that all samples in this dataset belong to honest prosumers.

\begin{figure}[t]
    \centering
    \begin{subfigure}[b]{0.95\linewidth}
        \centering
        \includegraphics[width=\linewidth]{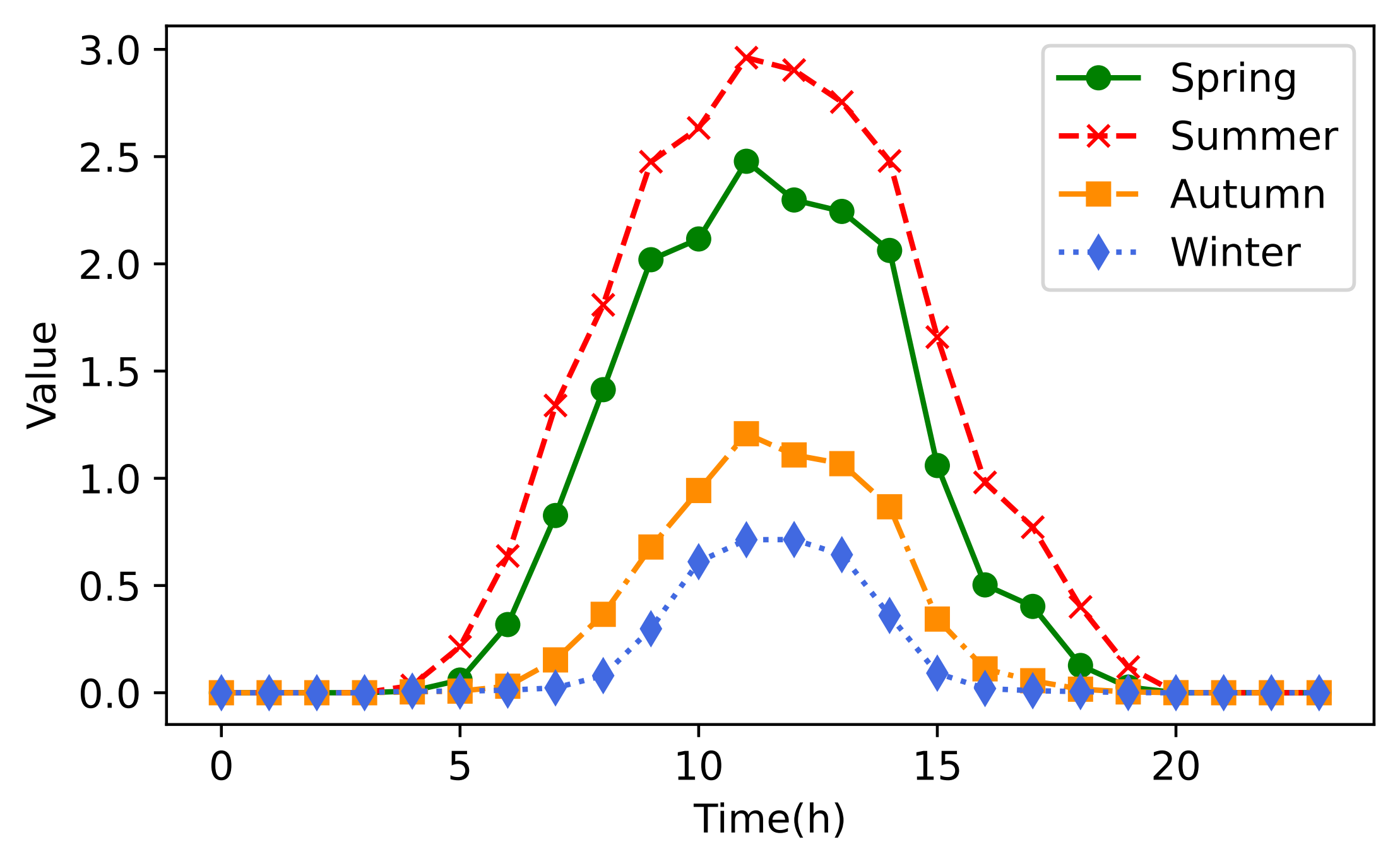}
        \caption{Seasonal Patterns of PV Generation.}
        \label{fig:data_norm}
    \end{subfigure}
    
    \begin{subfigure}[b]{0.95\linewidth}
        \centering
        \includegraphics[width=\linewidth]{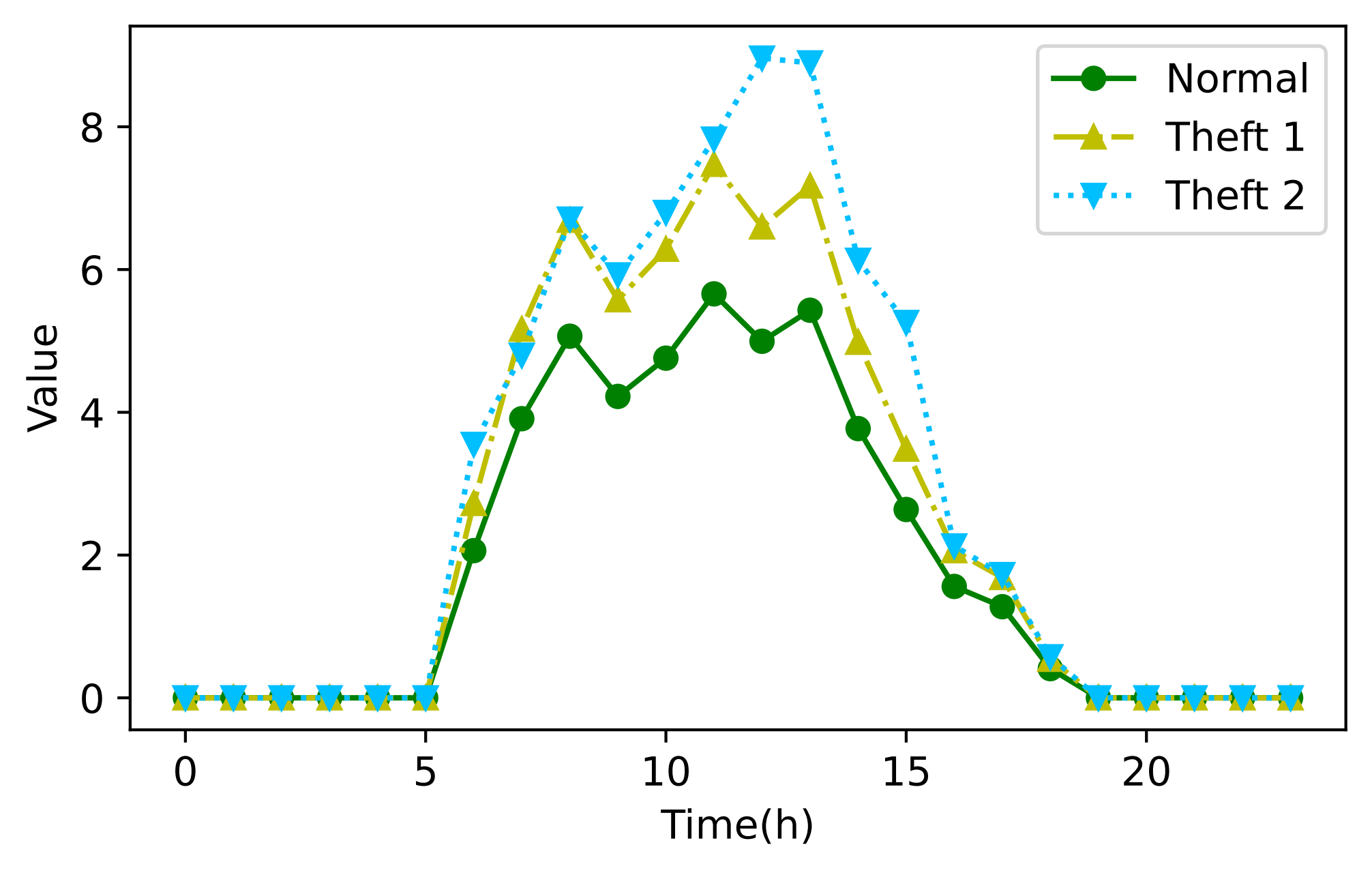}
        \caption{A customer's daily generation alongside the corresponding theft patterns.}
        \label{fig:data_theft}
    \end{subfigure}
    \caption{Patterns of PV generation.}
\end{figure}

Due to the absence of publicly available theft data in the generation domain, we simulate malicious samples based on prior research by applying cyber-attack functions~\cite{8998142} to benign samples. We employ two distinct types of cyber-attack functions to alter the daily generation patterns. The first type of theft involves increasing the daily reported electricity generation by a fixed percentage, while the second type randomly inflates the electricity generation at each reading. In our synthetic dataset, benign samples constitute 50\% of the data, while malicious samples generated from the first type of theft represent 25\%, and those from the second type of theft also account for 25\%. Suppose the daily electricity generation vector $E$ is given as $E = \{e_{1}, e_{2}, ..., {e_{24}}\}$ where $x_{t}$ represents the hour production for $t = 1,2,...,24$. Thus, two types of theft can be formulated as 
\begin{align}
    \text{Theft}_1(e_t)=(1+\alpha)e_t, \ \ \alpha=\text{random}(0.1,0.8), \\
\text{Theft}_2(e_t)=(1+\beta_t)e_t, \ \ \beta_t=\text{random}(0.1,0.8).
\end{align}
Figure~\ref{fig:data_norm} illustrates the daily variation trends of PV generation for distributed solar customers across four seasons in the dataset. The highest generation is observed in summer, followed by spring. Autumn and winter have noticeably lower generation, with winter showing the weakest overall generation throughout the day. This reflects the seasonal characteristics of PV generation, where summer has the longest daylight hours and the highest generation, while winter has the least sunlight and the lowest generation.
Additionally, Figure~\ref{fig:data_theft} presents an example of a customer's daily generation alongside the corresponding theft patterns. Under the theft patterns, the customer's reported generation is higher than the actual generation, with the discrepancy being particularly noticeable during midday when sunlight is abundant.

\subsection{Metrics and Benchmarks for Comparison}
To comprehensively measure the performance of the model, we adopt multiple evaluation metrics, including Accuracy, F1-Score, and Area Under the Curve (AUC).

Two types of benchmarks are selected for comparison with our proposed method: machine learning models and deep learning models. For the machine learning models, we employ classical algorithms such as Support Vector Classifier (SVC)~\cite{platt1998sequential}, Random Forest (RF)~\cite{breiman2001random}, and shape Dynamic Time Warping (shapeDTW)~\cite{zhao2018shapedtw}. In contrast, the deep learning models include CNN, Multi-scale Attention CNN (MACNN)~\cite{chen2021multi}, LSTM~\cite{graves2012long}, and Transformer~\cite{vaswani2017attention}. By comparing the performance of these two categories of models with our proposed method, we aim to comprehensively evaluate the effectiveness of various methods in addressing the problem.

\subsection{Result Analysis}

\begin{table}[]
\centering
\setlength{\tabcolsep}{9pt}
\caption{Comparative test results.}
\label{table:result}
\begin{tabular}{ccccc}
\hline\hline
\textbf{Category} & \textbf{Method} & \textbf{Acc.(\%)} & \textbf{F1(\%)} & \textbf{AUC(\%)} \\ \hline
\multirow{3}{*}{ML} & SVC             & 79.01             & 73.29           & 82.45            \\
&RF              & 76.23             & 69.23           & 83.32            \\
&ShapeDTW        & 84.33             & 82.54           & 84.00            \\ \hline
\multirow{5}{*}{DL} & CNN             & 89.29             & 89.60           & 92.11            \\
&MACNN           & 91.79             & 91.61           & \textbf{95.77}   \\
&LSTM            & 93.06             & 92.93           & 92.39            \\
&Transformer     & 92.07             & 91.99           & 90.86            \\ \cline{2-5}
&Proposed        & \textbf{94.57}    & \textbf{94.73}  & 93.81            \\
\hline\hline
\end{tabular}
\end{table}

To eliminate the effect of model performance randomness in the fair comparison, the experiments are repeated five times and the average results are presented. The detection performance in Table~\ref{table:result} provide a comprehensive comparison of different methods. The proposed method outperforms all other models across most evaluation metrics, achieving the highest accuracy and F1-score, along with a strong AUC. This indicates that the proposed method not only correctly identifies a high percentage of instances but also balances precision and recall effectively.

\begin{figure}[t]
    \centering
    \begin{subfigure}[b]{0.49\linewidth}
        \centering
        \includegraphics[width=\linewidth]{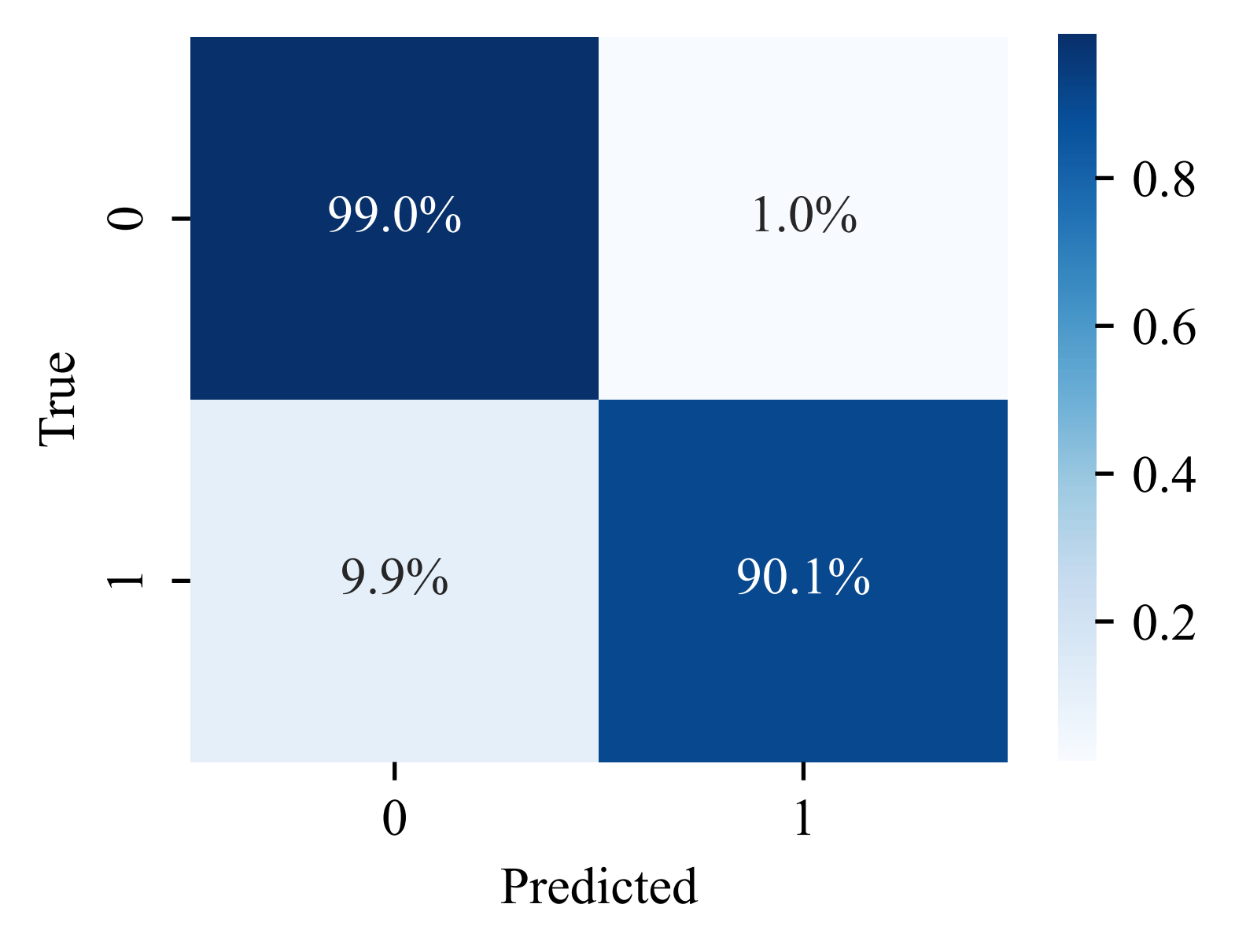}
        \caption{Proposed}
        \label{fig:Proposed_m}
    \end{subfigure}
    \begin{subfigure}[b]{0.49\linewidth}
        \centering
        \includegraphics[width=\linewidth]{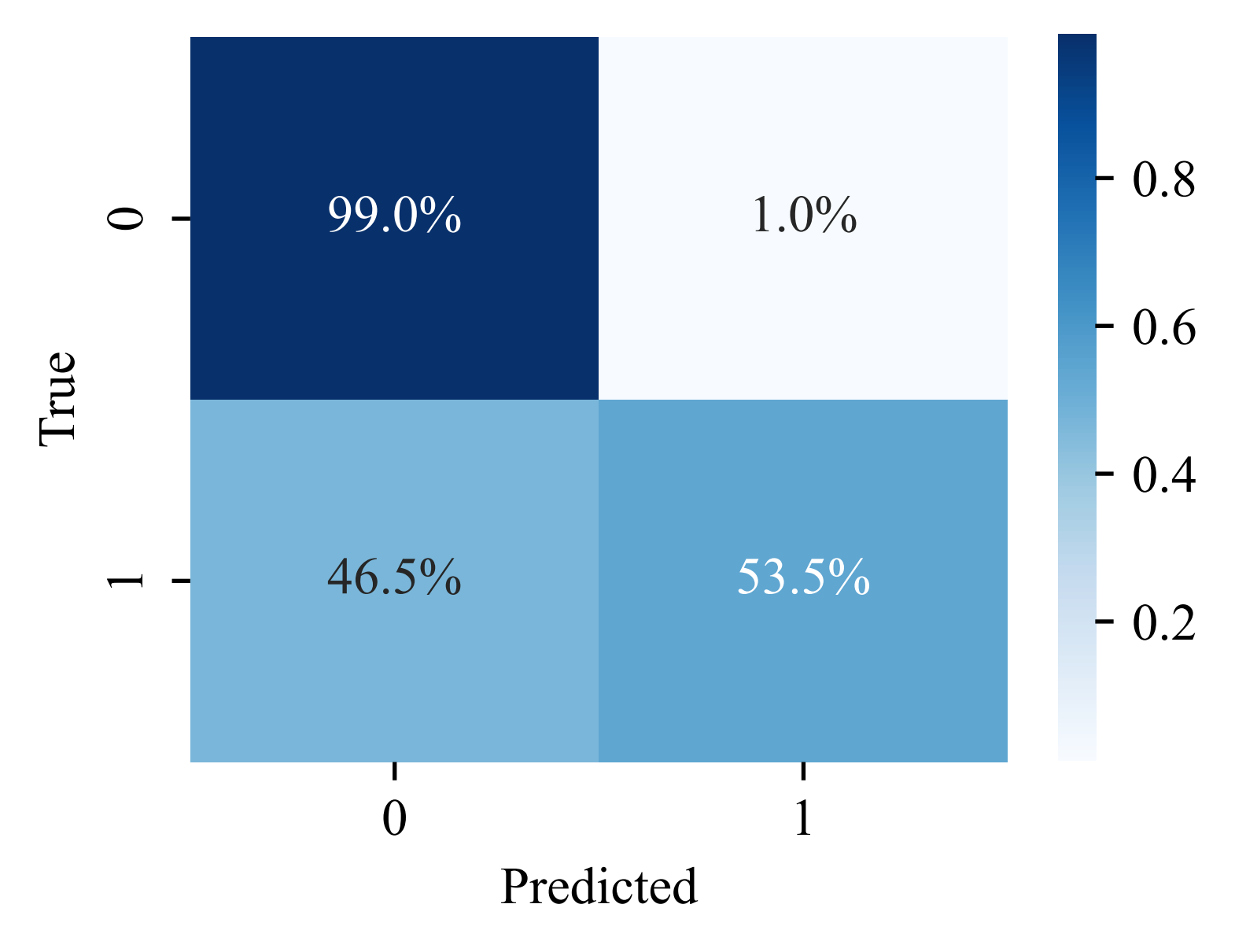}
        \caption{RF}
        \label{fig:RF}
    \end{subfigure}
    
   \vspace{0.05\linewidth}
    
    \begin{subfigure}[b]{0.49\linewidth}
        \centering
        \includegraphics[width=\linewidth]{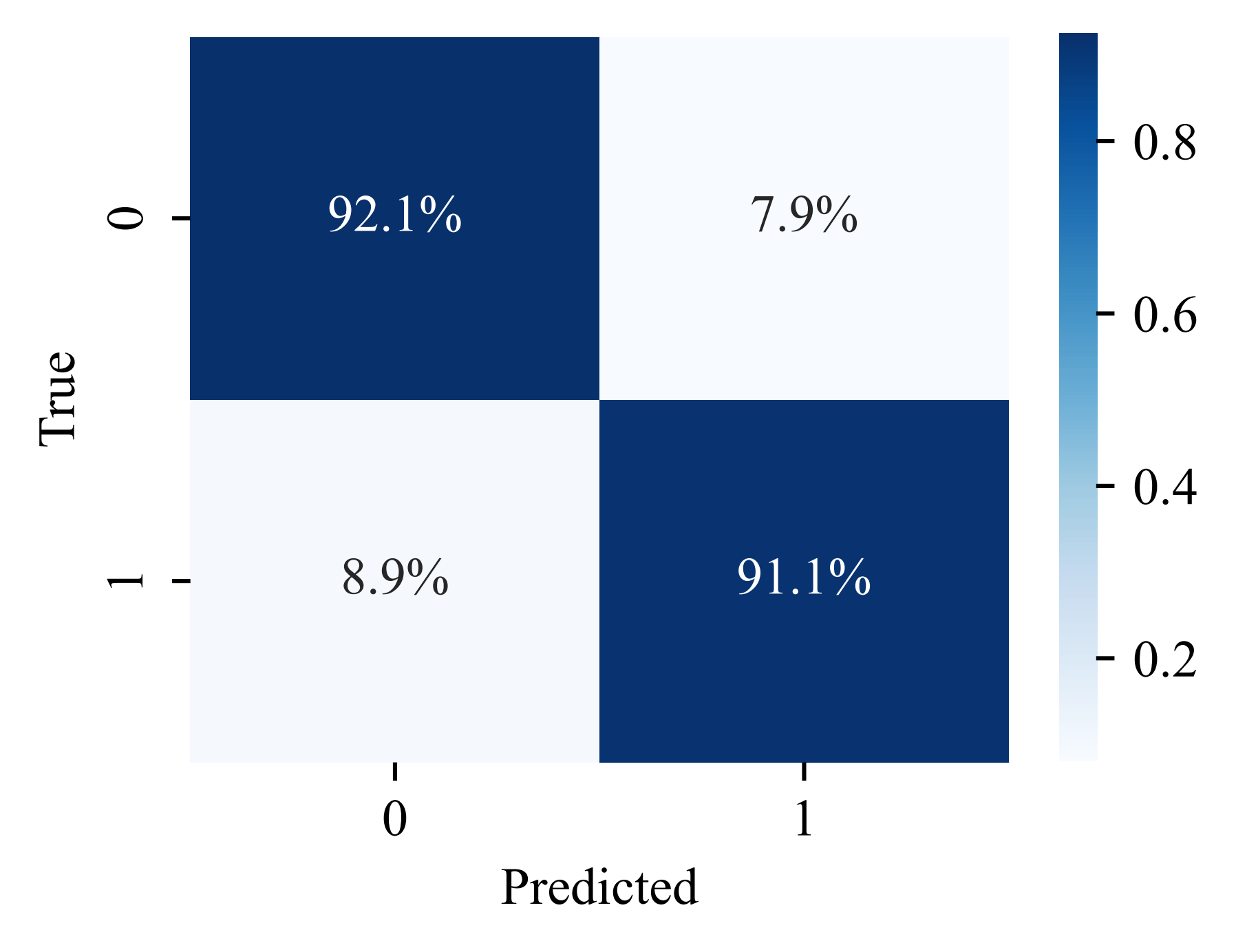}
        \caption{MACNN}
        \label{fig:MACNN}
    \end{subfigure}
    \begin{subfigure}[b]{0.49\linewidth}
        \centering
        \includegraphics[width=\linewidth]{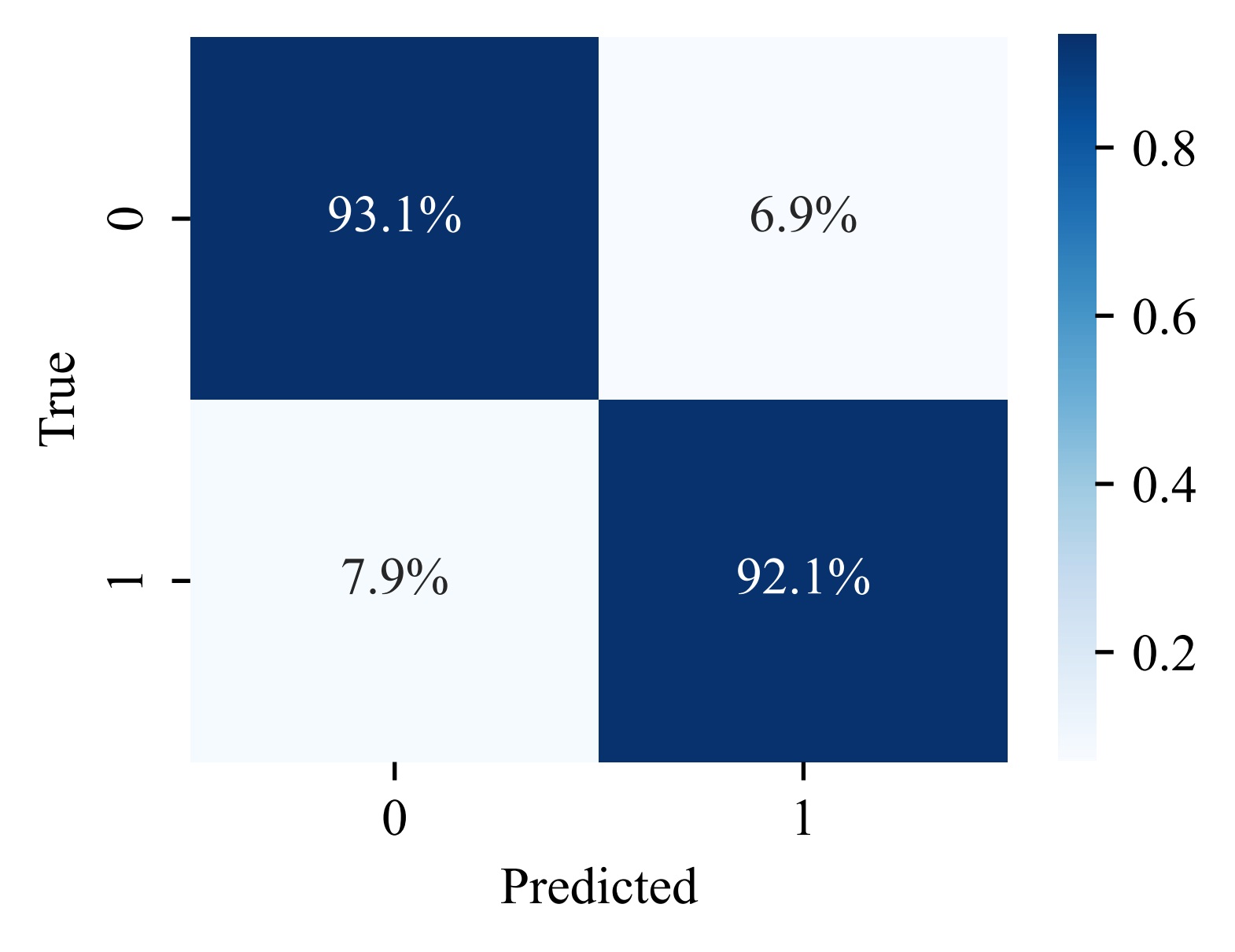}
        \caption{Transformer}
        \label{fig:Transformer}
    \end{subfigure}
    \caption{Confusion matrices of the proposed method, RF, MACNN, and Transformer.}
    \label{fig:cm}
\end{figure}

Among the deep learning models, MACNN stands out with an AUC of 95.77\%, the highest among all methods. This suggests that MACNN is particularly effective in distinguishing between positive and negative classes across different decision thresholds, although it slightly trails the proposed method in accuracy and F1-score. The CNN, LSTM, and Transformer also show strong performance, with LSTM achieving an accuracy of 93.06\% and an F1-score of 92.93\%. 
On the other hand, the traditional machine learning models like SVC and RF show lower performance compared to the deep learning models, indicating potential weaknesses in handling complex data patterns. ShapeDTW, another traditional method, performs moderately, outperforming RF and SVC but still lagging behind deep learning models. In summary, the proposed method demonstrates clear advantages over both traditional and deep learning-based benchmark models, particularly in terms of accuracy and F1-score.

The confusion matrices for several methods are presented in Figure~\ref{fig:cm}. The proposed method clearly outperforms the other models, particularly in its ability to accurately detect both true positives and true negatives, with the lowest rates of false positives and false negatives. 
Our method demonstrates strong performance in detecting benign samples, however, its accuracy diminishes when detecting malicious samples. The Random Forest model, in particular, struggles with predicting malicious samples, which may be attributed to the inherent limitations of traditional machine learning techniques. These limitations could include the model's inability to capture complex patterns and relationships in the data that are crucial for accurately identifying malicious samples.

\section{Conclusion}
In this study, we proposed a hybrid DL model that integrates multi-scale CNN, LSTM, and Transformer to effectively detect energy theft in residential PV generation systems. Our approach excels in capturing complex temporal dependencies and seamlessly integrates multi-source data, including temperature statistics, to improve detection accuracy. Our method was validated through extensive simulations using real-world data, demonstrating its robustness and effectiveness in maintaining the stability and fairness of energy systems in smart cities.

Future work will focus on enhancing the model's scalability and applying it to larger and more diverse datasets. Additionally, exploring ETD with both fraudulent load and PV generation is of great value for further research.

\bibliographystyle{IEEEtran}
\bibliography{references.bib}

\end{document}